\documentclass[mnsc,blindrev]{informs3clean}

\SingleSpacedXI

\usepackage{array}
\usepackage{url}
\usepackage{xparse}
\usepackage{subcaption}
\usepackage[margin=1in]{geometry}

\usepackage{todonotes}
\usepackage{bm}
\usepackage{epsfig}
\usepackage{float}
\usepackage{graphicx}
\usepackage{epstopdf}
\usepackage{amsfonts}
\usepackage{comment}
\usepackage{algorithm}
\usepackage{algpseudocode}
\usepackage{enumerate}
\usepackage{subcaption}
\usepackage{multirow}

\usepackage{natbib}
 \bibpunct[, ]{(}{)}{,}{a}{}{,}%
 \def\bibfont{\small}%

\newcommand{\overbar}[1]{\mkern 1.5mu\overline{\mkern-1.5mu#1\mkern-1.5mu}\mkern 1.5mu}

\TheoremsNumberedThrough     
\ECRepeatTheorems

\EquationsNumberedThrough    

\MANUSCRIPTNO{MS-0001-1922.65}

\begin{document}



\RUNTITLE{Trustworthy Feature Importance Avoids Unrestricted Permutations}

\TITLE{Trustworthy Feature Importance Avoids Unrestricted Permutations}


\ARTICLEAUTHORS{%
\AUTHOR{Emanuele Borgonovo}
\AFF{Department of Decision Sciences and Bocconi Institute for Data Science and Analytics, Bocconi University, Milan, Italy, \EMAIL{emanuele.borgonovo@unibocconi.it}} 
\AUTHOR{Francesco Cappelli}
\AFF{Department of Decision Science, Bocconi University, Milan, Italy, \EMAIL{francesco.cappelli@phd.unibocconi.it}}
\AUTHOR{Xuefei Lu}
\AFF{SKEMA Business School, Université Côte d’Azur, France, \EMAIL{xuefei.lu@skema.edu}} 
\AUTHOR{Elmar Plischke}
\AFF{Clausthal University of Technology, Clausthal-Zellerfeld, Germany, \EMAIL{elmar.plischke@tu-clausthal.de}}
\AUTHOR{Cynthia Rudin}
\AFF{Department of Computer Science, Duke University, Durham NC, 27708 USA, \EMAIL{cynthia@cs.duke.edu}}
} 

\ABSTRACT{
Feature importance methods using unrestricted permutations are flawed due to extrapolation errors; such errors appear in all non-trivial variable importance approaches. We propose three new approaches: conditional model reliance and  Knockoffs with Gaussian transformation, and restricted ALE plot designs. Theoretical and numerical results show our strategies reduce/eliminate extrapolation.
\color{black}
}%

 \HISTORY{This paper was
first submitted on April 12, 1922 and has been with the authors for
83 years for 65 revisions.}

\maketitle

%

\textbf{This version was submitted to the 2024 INFORMS Data Mining Best Paper Competition Awards (see \url{https://connect.informs.org/data-mining/awards/new-item22275866739965778}.)}

\section{Introduction.}\label{sec:intro} 
Since their introduction by \cite{Brei02}, permutation-based feature importance measures have been widely adopted. However, randomly permuting the entries of a dataset may create new points far from the original data or even \textquotedblleft impossible data.\textquotedblright{} In a permuted dataset, we may find children who are retired or individuals who graduated from high school before they were born \citep[p. 1]{MaseOwen22}. Forcing ML models to make predictions at these points causes them to extrapolate, making explanations unreliable \citep{Hooker2021}. Every non-trivial permutation-based variable importance measure, including SHAP \citep{LundLee17}, Knockoffs \citep{BarbCand15}, conditional model reliance \citep{FishRudi19}, and accumulated local effect (ALE) plots \citep{Apley2020} suffer from this. 
\color{black}

We propose and compare three new strategies to address extrapolation issues. The first combines conditional model reliance from \cite{FishRudi19} with a Gaussian transformation. By mapping data quantiles to a Gaussian distribution and back, we adjust only the quantiles of point values, significantly reducing extrapolation. Under a Gaussian copula assumption for the feature distribution, we prove that the new data points follow the same probability distribution as the original data. The second strategy applies the Gaussian transformation method to Knockoffs, improving over the approach in \cite{Hooker2021}, which uses Knockoffs to restrict permutations in Breiman's variable importance measures. We first applying a Gaussian transformation, and after permutation using Knockoffs with the Gaussian transformation, we map the new points back to the original space.
The third strategy combines Apley and Zhu's ALE plot design with Jansen's estimator \citep{Jansen_cpc_1999} to create a total-effect-like index. To address numerical noise from a refined ALE grid, we introduce a second index that averages squared Newton ratios. We also derive the theoretical relationship between permutation-based importance measures and total indices, with and without permutation restrictions. Our new variable importance metrics show a dramatic improvement over their unrestricted counterparts in being able to correctly identify the important variables for the data generation process.
\color{black}

\color{red}
\color{black}

\section{Background Literature \label{sec:LitRev}}
\paragraph{2.1 Breiman's Variable Importance Measures. \label{Sec_TIandALE}}
In the context of \cite{Hastie19941255}, analysts have a dataset
of feature and target realizations and aim to determine the relationship 
\begin{equation}
\mathbf{Y}=g(\mathbf{X},\mathcal{E)}\text{,}  \label{eq:ygxE}
\end{equation}%
where $\mathbf{X}$, $\mathbf{Y}$ and $\mathcal{E}$ are regarded as random
variables on a probability space $(\Omega ,\mathcal{B}(\Omega ),\mathbb{P})$%
, with $X\in \mathcal{X}$, $\mathcal{X}\subseteq \mathbb{R}^{d}$, $g:%
\mathcal{X}\rightarrow \mathbb{R}^{m}$, $\mathcal{E}:\mathcal{X}\times
\Omega \rightarrow \mathbb{R}$. Throughout the work, we suppose that $%
\mathcal{X}=\mathcal{X}_{1}\times \mathcal{X}_{2}\dots \times \mathcal{X}%
_{n} $, where $\mathcal{X}_{j}$ is the support of $X_{j}$, $j=1,2,...,d$. We
denote the vector of features excluding $X_{j}$ and the corresponding
support by $\mathbf{X}_{-j}$ and $\mathcal{X}_{-j}=\mathcal{X}\setminus 
\mathcal{X}_{j}$, respectively.

The map $g(\mathbf{X},\mathcal{E)}$ is assumed to be unknown and is
approximated by a model $\widehat{g}(\mathbf{x};\theta )$, $\widehat{g}:%
\mathcal{X}\times \Theta \rightarrow \mathbb{R}^{m}$, where $\theta \in
\Theta $ is a set of (hyper) parameters or rules. The parameters
determine $\widehat{g}$ via the solution of the (population version of the)
optimization problem: 
\begin{equation}
\min_{\theta \in \Theta }\mathbb{E}[\mathcal{L}(Y,\widehat{g}(\mathbf{X}
;\theta ))]\text{,}  \label{eq:MLOpt}
\end{equation}%
where $\mathcal{L}:\mathbb{R}^{m}\times \mathbb{R}^{m}\rightarrow \lbrack
0,\infty )$ is a loss function, with $\mathcal{L}(a,a^{\prime })=0$ if $%
a=a^{\prime }$ for all $a,a^{\prime }\in \mathbb{R}^{m}$. 
In practice, for a
dataset $D=\{(\mathbf{x}^{n},\mathbf{y}^{n});n=1,2,\dots ,N\}$ containing $N$
realizations of $(\mathbf{X},\mathbf{Y})$, the sample version of Problem %
\eqref{eq:MLOpt} aims to find $\theta ^{\ast }=\arg \, \min \{ \frac{1}{N}%
\sum_{n=1}^{N}\mathcal{L}(\mathbf{y}^{n},\widehat{g}(\mathbf{x}^{n};\theta
))\}$ with $(\mathbf{x}^{n},\mathbf{y}^{n})\in D$.
We use the simplified notation $\widehat{g}(\mathbf{X})$ instead of  $\widehat{g}%
(\mathbf{X};\theta ^{\ast })$ for the trained ML model henceforth.
\color{black}

The (empirical) permutation feature importance of $X_{j}$ under model $%
\widehat{g}(\mathbf{\cdot };\theta ^{\ast })$ is defined by 
\begin{equation}
\widehat{\nu }_{j}=\frac{1}{N}\sum_{n=1}^{N}\mathcal{L}\left( \mathbf{y}^{n},%
\widehat{g}(\mathbf{x}_{j,\text{perm}}^{n};\theta ^{\ast })\right) -\frac{1}{%
N}\sum_{n=1}^{N}\mathcal{L}\left( \mathbf{y}^{n},\widehat{g}(\mathbf{x}%
^{n};\theta ^{\ast })\right) ,  \label{eq:nuhat1}
\end{equation}%
where $\frac{1}{N}\sum_{n=1}^{N}\mathcal{L}(\mathbf{y}^{n},\widehat{g}(%
\mathbf{x}^{n};\theta ^{\ast }))$ is the expected minimal loss for the
sample and $\frac{1}{N}\sum_{n=1}^{N}\mathcal{L}(\mathbf{y}^{n},\widehat{g}(%
\mathbf{x}_{j,\text{perm}}^{n};\theta ^{\ast }))$ is the average loss when
we randomly permute the entries of $X_{j}$. 
The method is model-agnostic, but the same feature $X_{j}$ can be assigned different values of $\widehat{\nu }_{j}$ depending on the chosen model $\widehat{g}$. To address this, \cite{FishRudi19} introduce the concept of model class reliance, considering $\widehat{\nu }_{j}$ across all near-optimal predictive models, known as the Rashomon set \citep[see][]{FishRudi19,SemeRudi22}. 
\color{black}

\paragraph{2.2 Total Indices\label{sec:globalsensitivity}.}
Total indices (${\tau }_{j}$) are feature importance measures that originate in the
simulation literature \citep{Homma1996,saltelli_jasa_2002} and have
attracted recent attention for explainability 
\citep{HartGrem18,
BenaDaVe22,HuanRosh24}. The total index of feature $X_{j}$ is
the expected portion of the variance of $Y$ that remains after all features
are fixed but $X_{j}$.  
\color{gray}
\color{black}
The works of \cite{Jansen_cpc_1999} and  \cite{KuchTara12} show that ${\tau }_{j}$ can be written as 
\begin{equation}
{\tau }_{j}=\frac{1}{2}\left( \mathbb{E}\left[ \left( g(X_{j}^{\prime },%
\mathbf{X}_{-j})-g\left( \mathbf{X}\right) \right) ^{2}\right] \right) ,
\label{eq:total_index}
\end{equation}%
where $\mathbf{X}^{\prime }$ is an independent replicate
of $\mathbf{X}$. 
When
features are correlated, Equation \eqref{eq:total_index} is equivalent to: 
\begin{equation}
\tau _{j}=\frac{1}{2}\int_{\mathcal{X}}\int_{\mathcal{X}_{j}}\left( g\left(
x_{j}^{\prime },\mathbf{x}_{-j}\right) -g\left( \mathbf{x}\right) \right)
^{2}dF_{X_{j}|\mathbf{X}_{-j}}(x_{j}^{\prime }|\mathbf{x}_{-j})dF_{\mathbf{X}%
}(\mathbf{x}),  \label{eq:tauicorr}
\end{equation}%
where $F_{X_{j}|\mathbf{X}_{-j}}(x_{j}^{\prime }|\mathbf{x}_{-j})$ is the
conditional distribution of $X_{j}$ given $\mathbf{X}_{-j}$, and $F_{\mathbf{%
X}}(\mathbf{x})$ the joint distribution of $\mathbf{X}$.
When $X_{j}^{\prime }$ is obtained from a free permutation, i.e., sampled
from the marginal distribution of $X_{j}$, Equation \eqref{eq:total_index}
becomes 
\begin{equation}
\tau _{j}^{\prime }=\frac{1}{2}\int_{\mathcal{X}}\int_{\mathcal{X}%
_{j}}\left( g\left( x_{j}^{\prime }, \mathbf{x}_{-j}\right) -g\left( \mathbf{x%
}\right) \right) ^{2}dF_{X_{j}}(x_{j}^{\prime })dF_{\mathbf{X}}(\mathbf{x})%
\text{.}  \label{eq:tauiprimeLoco}
\end{equation}
Clearly, under feature
independence, $\tau _{j}^{\prime }$ is the total index of $X_{j}$ for $%
\widehat{g}(\mathbf{X})$. It can be proven that $\tau _{j}^{\prime }=0$ if
and only if $\widehat{g}(\mathbf{X})$ does not depend on $X_{j}$ \citep
{{VerdWass23}}.

\paragraph{2.3 ALE Plots.}
\cite{Apley2020} propose accumulated local effects (ALE) plots as a new visualization tool that reduces the risk of extrapolation. Assuming
that $\widehat{g}(\cdot )$ is differentiable, the ALE function of $X_{j}$ is
defined as 
\begin{equation}
ALE_{j}(x_{j})=\int_{x_{j,\text{min}}}^{x_{j}}\mathbb{E}_{\mathbf{X}_{-j}|X{_{j}}}[ 
\widehat{g}_{j}^{\prime }(\mathbf{X})|X_{j}=t_{j}]d{t_{j}}\text{,}
\label{MR=ale}
\end{equation}%
where $\widehat{g}_{j}^{\prime }(\mathbf{X})$ is the partial derivative of $%
\widehat{g}$ with respect to $X_{j}$ and $x_{j,\text{min}}$ is a chosen value close
to the lower bound of the support of the distribution of $X_{j}$. 
\paragraph{2.4 Model-X Knockoffs\label{Sec:Knock}}
Model-X Knockoffs, introduced by \cite{BarbCand15} and \cite{Candes2018}, are a feature selection method designed to control the false discovery rate by generating artificial variables, called knockoffs. These knockoffs replicate the original variables and are used to create test statistics to assess the relevance of the original variables. Under certain conditions, knockoff variables follow the same distribution as the original variables, a property exploited by \cite{Hooker2021} to limit the impact of permutations in their tests -- but not in most realistic cases. In fact, in all of these approaches, and in other approaches such as SHAP \citep{LundLee17}, which is too expensive to use for large scale problems anyway, permutations are done without regard to the possibility of extrapolation.
\paragraph{2.5: Shapley Additive Explanations \label{Sec:SHAPs}}
Shapley values are widely studied in computer science and artificial
intelligence as feature importance measures ---  \cite{SundNajim20}. \cite{LundLee17} provide a unified perspective and a computational framework aimed at efficiently estimating SHapley Additive exPlanations (SHAPs). \citet{LundLee17}
perform an in-depth analysis, and to reduce comptational burden suggest
replacing the retraining with averaging while fixing all possible feature
subsets. As underlined by \cite{MaseOwen22}, this strategy involves the same
operations within the calculation of partial dependence functions and
is, therefore, exposed to extrapolation issues. Thus, calculation of the
SHAPs is exposed to the same extrapolation issues as partial
dependence functions. However, to our knowledge, a systematic investigation
of the intensity with which this effect occurs has not previously been performed. We address this point in our experiments.

\color{black}
\section{Three New Strategies}
\paragraph{3.1: Conditional Model Reliance after Gaussian Transformation}\label{sec:GCMR}
Our first approach is inspired by the residual imputation strategy and by
the notion of conditional model reliance of \citet{FishRudi19}. The intuition is to randomly permute the residuals instead of the original points. The conditional model reliance calculation starts with regressing the feature of interest ($X_j$) on the remaining features ($\mathbf{X}_{-j}$) via $g_{impute}(\mathbf{X}_{-j})=\mathbb{E}[X_j|\mathbf{X}_{-j}]$. Then, randomly permute the residuals $\overbar{X}_j:=X_j-\mathbb{E}[X_j|\mathbf{X}_{-j}]$ to obtain $\overbar{X}_j^{\pi}$. A new point $X_j^\prime $ is then defined as as $X_j^\prime:=\mathbb{E}[X_j|\mathbf{X}_{-j}]+\overbar{X}_j^{\pi}$.  From there, the total index, as the difference between the imputed data's loss and the original data's loss, as in \eqref{eq:total_index}, can be used to assess the variable importance of the unique information carried by variable $j$ that is not included in other variables.
However, even with permuting only residuals,
it constructs points that are out of distribution, see Figure~\ref{g_impute}(b). 
\color{black}

Our fix for this problem is to perform residual imputation \textit{after
transforming the data into normal scores}. The procedure can be summarized in Algorithm~\ref{algoGauss}. 
\begin{algorithm}[H]
	\caption{Steps of the Gaussian-transformation and conditional model reliance strategy (GCMR).}\label{algoGauss}
	\begin{flushleft}
		\textbf{Input:} feature dataset $\widehat{X}$. \hfill \\
		\textbf{Output:} A permuted dataset $\widehat{X}^\prime$ that avoids unrestricted permutations.
	\end{flushleft}
	\begin{algorithmic}
		\For{$j=1,2,\dots,d$}
		\State Standardize $\widehat{X}_j$ 
		\State Map the quantiles of $\widehat{X}_j$ into the quantiles of a standard normal random variable $Z_j$
		\EndFor
		\For{$j=1,2,\dots,d$}
		\State Calculate the non-parametric regression curve $\mathbb{E}[Z_j|\mathbf{Z}_{-j}] $
		\State Calculate the residual $\overbar{Z}_j=Z_j-\mathbb{E}[Z_j|\mathbf{Z}_{-j}] $
		\State Permute $\overbar{Z}_j$ to obtain $\bar{Z}_j^{\pi}$ 
		\State Add back $\overbar{Z}_j^{\pi}$ to $\mathbb{E}[Z_j|\mathbf{Z}_{-j}]$ to obtain $Z_j^\prime=\mathbb{E}[Z_j|\mathbf{Z}_{-j}] + \overbar{Z}_j^{\pi}$
		\State Map $Z_j^\prime$ into $X_j^\prime$
		\EndFor     
	\end{algorithmic}
\end{algorithm}

Note that in the second \textit{for} loop of Algorithm~\ref{algoGauss}, we edit only z-scores, which is identical to adjusting the
quantile of $X_j$ within its marginal distribution, which does not go out of
range. For instance, even if $Z^{\prime }_j$ is set to the lowest quantile,
it would still correspond to the lowest values of $X_j$ that are realized in
the dataset.
\color{black}

\begin{example} \label{example:testcase3}
In this example, we consider two uniformly distributed random variables ($X_1$
and $X_2$) on the interval $[0, 1]^2$. We assume that $X_1$ and $X_2$ are
strongly correlated ($\rho=0.95$). We model the correlation through a
Gaussian copula. In Figure \ref{g_impute}, we report original data (black)
and new data (red) obtained, respectively, after an unrestricted permutation
of $X_1$ (Left), a residual imputation (Middle) and a residual
imputation with Gaussian transformation (Right). 
\begin{figure}[h!]
\centering
\begin{minipage}{0.32\linewidth}
		\includegraphics[width=\textwidth]{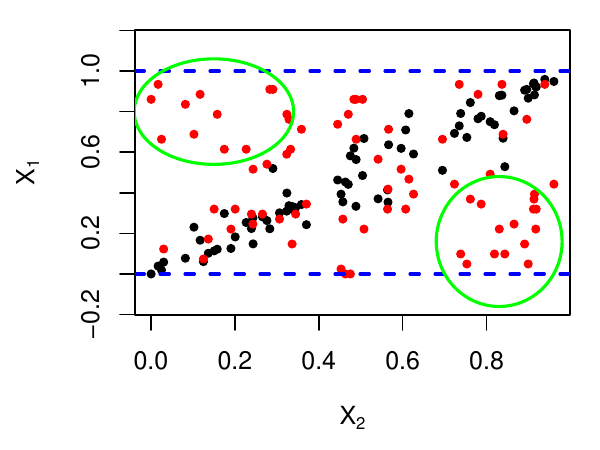}
	\end{minipage} 
\begin{minipage}{0.32\linewidth}
		\includegraphics[width=\textwidth]{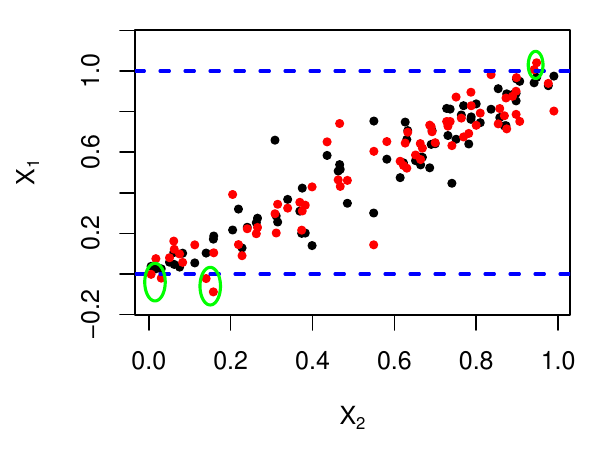}
	\end{minipage} 
\begin{minipage}{0.32\linewidth}
		\includegraphics[width=\textwidth]{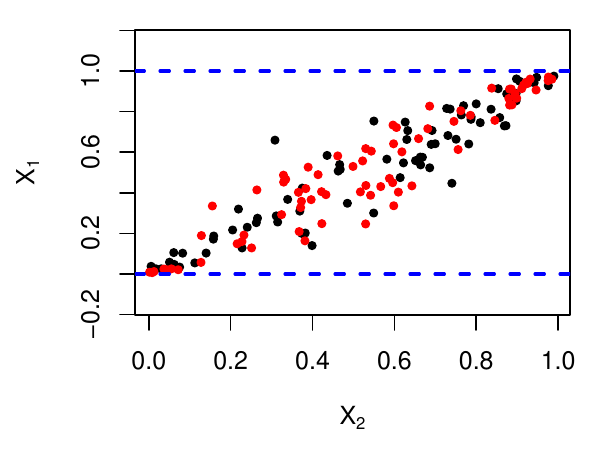}
	\end{minipage}
\caption{Original (black) and permuted data (red). (Left) with unrestricted
permutations; (Middle) with residual imputation without Gaussian transformation;
(Right) with residual imputation with Gaussian transformation (our proposed approach).}
\label{g_impute}
\end{figure}
Figure \ref{g_impute} (Left) shows that unrestricted permutation may result in new data points far from the original ones (highlighted with a green oval), necessitating extrapolation by an ML model. Figure \ref{g_impute} (Middle) shows that the permutation restricted through residual imputation results in new points (red) closer to the original points (black), however, some still fall outside the support. Figure \ref{g_impute} (Right) illustrates a restricted permutation using residual imputation with a Gaussian transformation, where new points (red) remain within the cloud of original points and never exceed the support.
Furthermore, these new points preserve the original data dependence structure. 
\end{example}  
Similarly to \citet{FishRudi19}, calculations from
these new points would yield conclusions about the importance of the
unique information about the feature, i.e., the information this feature
contains that cannot be gleaned from other variables. We denote the indices by $\nu_j^\text{GCMR}$, to distinguish them from the unrestricted  $\nu_j$. 
On the theoretical side, we can prove the following (All proofs are in the full version).
\begin{proposition}
\label{prop:GaussCopula} If the joint distribution of the features $\mathbf{X%
}$ can be modeled by a Gaussian copula, then $\mathbf{X}^{\prime }$ returned
by Algorithm~\ref{algoGauss} is an independent copy of $\mathbf{X}$.
\end{proposition}
Proposition \ref{prop:GaussCopula} provides theoretical guarantees that the new
permuted points follow the same distribution as the original points when the
feature-generating process can be modeled by a Gaussian copula. 
\paragraph{3.2: The Gaussian Knockoff Strategy.}
\label{sec:GKnock} Our second approach explores the use of
knockoffs. We consider
mapping the original data into the corresponding Gaussian transformed random
variables, as in Algorithm \ref{algoGauss}. The
second \textit{for} loop is replaced by generating new values of $Z$ variables from
the knockoff procedure. The knockoffs $Z^\prime$ are then mapped back to the
original space, to form the permuted sample $X^\prime$. We call this
strategy GKnock. Similarly to the GCMR strategy, the following proposition
holds.
\begin{proposition}
\label{prop:GKnock} If the joint distribution of the features $\mathbf{X}$
can be modeled by a Gaussian copula, then $\mathbf{X}^{\prime }$ returned by
the Gknock procedure is a Knockoff copy of $\mathbf{X}$.
\end{proposition}
Applying GKnock to Example \ref{example:testcase3}
yields results similar to those in Figure \ref{g_impute} (not reported for the sake of space). The new points can be used to compute permutation-based importance measures with data that follow the variable marginal distributions. We use the symbol $\nu_j^\text{GKnock}$ to denote Breiman's importance measures when the new points are Knockoffs of the original variables. 

\begin{figure}[]
\centering
{\includegraphics[width=0.7\textwidth]{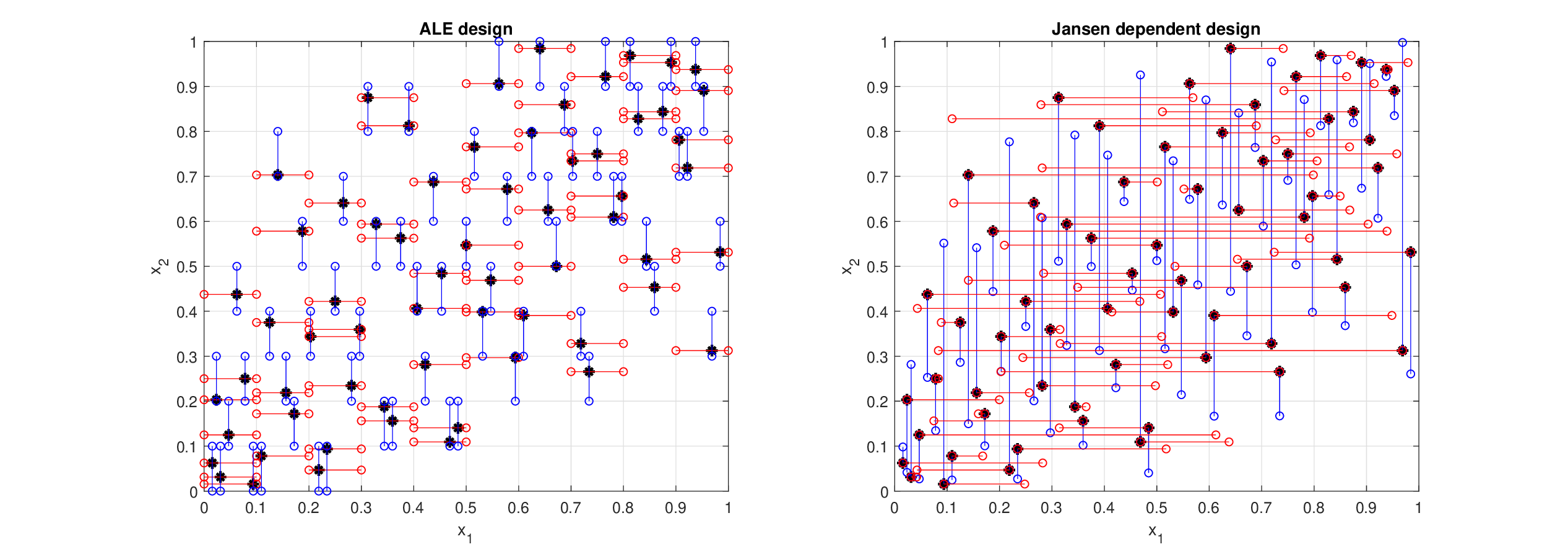}}
\caption{ALE design (left graph) and Jansen design (right graph) for a
correlated case. Legend: $\bullet $ original points; red (
\color{red}{$
\circ $}\color{black}{}), new points after varying $X_1 $;
blue ( \color{blue}{$ \circ $}\color{black}{}), new points after varying $%
X_2 $.}
\label{fig:AleDesign}
\end{figure}

\paragraph{3.3: The ALE Plots Feature Importance Measures. \label{sec:tauALE}}
The intuition of our third strategy strategy is to associate feature importance measures
with the design of ALE plots. As in \cite{Apley2020}, assume that $\mathcal{X}%
_{j}=[x_{j,\text{min}},x_{j,\text{max}}]$ is an interval on the real line
(or the union of a possibly disjoint set of intervals). We can partition $%
\mathcal{X}_{j}$ into $K$ subintervals $\mathcal{X}%
_{j}^{k}=[z_{j}^{k-1},z_{j}^{k})$, with $z_{j}^{0}=x_{j,\text{min}}$ and $%
z_{j}^{K}=x_{{j,\text{max}}}$, such that $\bigcup \limits_{k=1}^{K}\mathcal{X}%
_{j}^{k}=\mathcal{X}_{j}$ and $\mathcal{X}_{j}^{k}\bigcap \mathcal{X}%
_{j}^{m}=\emptyset $, $k,m=1,2,\dots ,K$, $k\neq m$.
). 
Then, as in \cite{Apley2020}, we denote with $n_{j}^{K}(k)$ the number of
realizations of $X_{j}$ that belong to $\mathcal{X}_{j}^{k}$, and with $%
k_{j}^{K}(x)$ the index of the subinterval that contains $x$. An estimate of 
$ALE_{j}(x_{j})$ is given by 
\begin{equation}
\widehat{ALE}_{j}(x_{j})=\sum_{k=1}^{k_{j}^{K}(x_{j})}\frac{1}{n_{j}^{K}(k)}%
\sum_{n:\mathbf{x}_{j}^{n}\in \mathcal{X}_{j}^{k}}\left( \widehat{g}\left(
z_{j}^{k},\mathbf{x}_{-j}^{n}\right) -\widehat{g}\left( z_{j}^{k-1},\mathbf{x%
}_{-j}^{n}\right) \right) \text{.}  \label{MR=ale3}
\end{equation}%
The calculation of $\widehat{ALE}_{j}(x_{j})$ requires averaging differences
in predictions over the conditional distribution of the feature of interest
and also for points close to a given realization $\mathbf{x}^{n}$. Note that
the $L^{1}$ distance between $\left( z_{j}^{k},\mathbf{x}_{-j}^{n}\right) $
and \ $\left( z_{j}^{k-1},\mathbf{x}_{-j}^{n}\right) $ is $%
|z_{j}^{k}-z_{j}^{k-1}|$. Supposing equally spaced partitions, we have $%
|z_{j}^{k}-z_{j}^{k-1}|=1/K$, so that the difference is bounded. Thus, the
new evaluation points are forced to lie close to the original points and
unrestricted permutations are avoided (Figure \ref{fig:AleDesign}).
The left graph of Figure \ref{fig:AleDesign} displays the points visited by
an ALE algorithm. Note that the new points $(z_{j}^{k},\mathbf{x}_{-j}^{k})$
(in light color) are always close
to the original data $\mathbf{x}^{k}$ (darker color) and this reduces
extrapolation issues. The right graph of Figure \ref{fig:AleDesign} shows
points visited by an algorithm in which $X_{j}^{\prime }$ is sampled
independently of $\mathbf{X}_{-j}$; the new points $(x_{j}^{^{\prime }},%
\mathbf{x}_{-j}^{k})$ can be far away from the original point $\mathbf{x}%
^{k} $, with potential extrapolation problems. 

We define two importance indices from ALE plots. The first is a total-ALE effect given by: 
\begin{equation}
\tau _{j}^\text{ALE}(K)=\dfrac{1}{2}\sum \limits_{k=1}^{K}\mathbb{E}[\left(%
\widehat{g}(z_{j}^{k},\mathbf{X}_{-j})-\widehat{g}(z_{j}^{k-1},\mathbf{X}%
_{-j})\right)^{2}|{X}_j\in \mathcal{X}_{j}^{k}]\cdot \mathbb{P}({X}_j\in 
\mathcal{X}_{j}^{k})\text{,}  \label{eq:tauALEcond}
\end{equation}%
where $\mathbb{P}({X}_j\in \mathcal{X}_{j}^{k})$ is the probability that the 
$j$th component of $\mathbf{X}$, $X_j$, belongs to the sub-interval $%
\mathcal{X}_{j}^{k}$. Equation \ref{eq:tauALEcond} makes it explicit the
dependence of $\tau _{j}^\text{ALE}(K)$ on the grid. Therefore, $\tau _{j}^\text{ALE}(K)\neq \tau _{j}^\prime$, because the ALE main effects $\Phi _{j}^{\text{ALE}}(X_j^\prime,\mathbf{X}_{-j};K)$ are
calculated for values of $X_j$ fixed at $z_k$ and $z_{k-1}$ for all
realizations of $\mathbf{X}_{-j}$. Conversely, the classical effects $\Phi_{j}^\prime(X_{j}^{\prime },\mathbf{X}
)$ are calculated with the new value $X_{j}^{\prime }$ sampled independently
from $\mathbf{X}_{-j}$. 
The second is a derivative-based index given by
\begin{equation}  \label{eq:kappa}
\kappa _{j}^\text{ALE}(K)=\frac{1}{K}\sum \limits_{k=0}^{K-1}\mathbb{E}\left[
\left( \frac{\widehat{g}(z_{j}^{k},\mathbf{X}_{-j})-\widehat{g} (z_{j}^{k-1};%
\mathbf{X}_{-j})}{z_{j}^{k}-z_{j}^{k-1}}\right) ^{2}\right] \frac{\sigma
_{j}^{2}}{\sigma _{y}^{2}}.
\end{equation}%
The index $\kappa _{j}^\text{ALE}(K) $ is a normalized expectation of Newton
ratios computed at randomized locations in the feature space in the spirit of
derivative-based sensitivity measures of \cite{sobol_mcs_2009} (see also \citealt{Song2019142}).  

\paragraph{4.4: Relating Permutation-based Importance Meausures and Total Indices.}

Let us consider the population definition of the variable importance
measures in Equation \eqref{eq:nuhat1}:%
\begin{equation}
\nu _{j}=\mathbb{E}[\mathcal{L}\left( Y,\widehat{g}(X_{j}^{^{\prime }},%
\mathbf{X}_{-j};\theta ^{\ast })\right) ]-\mathbb{E}[\mathcal{L}\left( Y,%
\widehat{g}(\mathbf{X};\theta ^{\ast })\right) ].
\label{eq:nujLosspopulation}
\end{equation}

\begin{proposition}
\label{prop:nutau}1) If $\mathcal{L}\left( \cdot ,\cdot \right) $ is the
quadratic loss and if the model is a perfect predictor, then%
\begin{equation}
\nu _{j}=\mathbb{E}[\left( \widehat{g}(\mathbf{X};\theta ^{\ast })-\widehat{g%
}(X_{j}^{^{\prime }},\mathbf{X}_{-j};\theta ^{\ast })\right) ^{2}].
\label{eq:nusquarepolulation}
\end{equation}%
Additionally, under the assumptions of item 1), \newline
2) If $X_{j}^{^{\prime }}$ is sampled independently of $\mathbf{X}_{-j}$,
then%
\begin{equation}
\nu _{j}=2\tau _{j}^{\prime },  \label{eq:nu2tauLoco}
\end{equation}%
where $\tau _{j}^{\prime }$ is the $\Psi _{DLoco}$ total index in Equation %
\eqref{eq:tauiprimeLoco}.\newline
3) If $X_{j}^{^{\prime }}$ is sampled conditionally on $\mathbf{X}_{-j}$,
then 
\begin{equation}
\nu _{j}=2\tau _{j},
\end{equation}%
where $\tau _{j}$ is the classical total index in Equation %
\eqref{eq:tauicorr}$.$
\end{proposition}

Proposition \ref{prop:nutau} sheds light on the meaning of Breiman's variable importance
measures with and without permutation restrictions. Without permutation
restrictions, Breiman's variable importance measures are equal to twice the $%
\Psi _{DLoco}$ importance of Verdinelli and Wassermann. With permutation restrictions, they are equal to twice the classical (Sobol') total indices.
For the empirical estimates $\widehat{\nu }_{j}$ in Equation %
\eqref{eq:nuhat1}, we have the following.
\begin{corollary}
\label{cor:estimates}Under the assumptions of Proposition \ref{prop:nutau}, 
\newline
1) $\widehat{\nu }_{j}$ in Equation \eqref{eq:nuhat1} is an asymptotically
unbiased estimator of $\nu _{j}$ in Equation \eqref{eq:nusquarepolulation}.%
\newline
2) If the new points $X_{j}^{^{\prime }}$ are obtained from a free
permutaiton of $X_{j}$, then%
\begin{equation}
\widehat{\nu }_{j}=2\widehat{\tau }_{j}^{\prime },
\end{equation}%
where $\tau _{j}^{\prime }$ is an estimate of the $\Psi _{DLoco}$ total
index in Equation \eqref{eq:tauiprimeLoco}.\newline
3) If, in addition, under the assumptions, respectively of Propositions \ref%
{prop:GaussCopula} or \ref{prop:GKnock}, if the new points $X_{j}^{^{\prime
}}$ are sampled through Algorithms $1$ or $2$, then 
\begin{equation}
\widehat{\nu }_{j}^{\text{GCMR}}=\widehat{\nu }_{j}^{\text{GKnock}}=2%
\widehat{\tau }_{j},
\end{equation}%
where $\tau _{j}$ is the classical total index in Equation %
\eqref{eq:tauicorr}.
\end{corollary}

\section{Numerical Experiments\label{NumericalExps}}
We tested the performance of the new strategies and compared them to traditional strategies in several experiments. We report some of the results here, while more experiments are described in the full paper.
\paragraph{4.1: Hooker's Analytical Test Case}
\cite{Hooker2021} design a case study to highlight the impact of extrapolation errors on Breiman's feature importance measures. The model is:
\begin{equation}
Y=g(X)=X_{1}+X_{2}+X_{3}+X_{4}+X_{5}+0\cdot X_{6}+0.5\cdot X_{7}+0.8\cdot
X_{8}+1.2\cdot X_{9}+1.5\cdot X_{10}+\epsilon,  \label{Hooker:model}
\end{equation}%
with $X_{j}\sim \mathcal{U}[0,1]$ and $\epsilon \sim \mathcal{N}(0,0.1^{2})$, so that the most important features are the ones associated with the coefficients of large magnitude. 
To test the effect of correlations, $X_{1}$ and $X_{2}$ are linked via a Gaussian copula with correlation coefficient $\rho_{X_1,X_2}$. 
Results of \cite{Hooker2021} show that the ranking induced by $\nu_j $ changes as $\rho_{X_1,X_2}$ varies. In particular, for high values of $\rho_{X_1,X_2}$, the machine learning models tend to attribute higher importance to $X_1$ and $X_2$.
\begin{table}[H]
    \centering
    \footnotesize
     \caption{Ground Truth values for $\nu_j=2\tau_j$ in the Hooker test case.}\label{tab:HookerTrue}
 \begin{tabular}{|l|c |c |c| c| c| c| c|}
 \hline
Feature & $X_1$ , $X_2$ & $X_3$, $X_4 $, $X_5$ & $X_6$ & $X_7$ & $X_8$ & $X_9$ & $X_{10}$  \\ \hline
$\rho_{X_1,X_2}=0.00$ &      0.1667     &      0.1667       &     0   &    0.0417    &  0.1067  &   0.2400 &  0.3750      \\ \hline
 $\rho_{X_1,X_2}=0.90$       &  0.0332 &  0.1756   &    0  &  0.0438 &   0.1124&    0.2528&    0.3950   \\ \hline
\end{tabular} 
\end{table}
In our experiments, we exploit two facts. First, when the true model $g$ is used, $\nu_j=2\tau_j$. This allows us to calculate the ground truth values of $\nu_j$ by using a subroutine that estimates total indices. The ground truth values are reported in Table \ref{tab:HookerTrue} for both the case $\rho(X_1,X_2)=0$ and $\rho(X_1,X_2)=0.9$. 
The values in Table \ref{tab:HookerTrue} show that, when introducing correlations, the ground truth value of  $\nu_j$ for $X_1$ and $X_2$ decreases with respect to the independent case.

Following \citet{Hooker2021}, we generate synthetic input-output data via Monte Carlo simulation with size $N=2000$ for each correlation assignment.  On each input-output dataset, we train a linear model (LM), a random forest (RF), and an artificial neural network (NN)  as in \citet{Hooker2021}, using the \textsc{Matlab} subroutines \textsc{fitlm}, \textsc{fitnet}, and \textsc{fitrensemble}. Then, we assess the models' out-of-sample performance on a newly generated dataset of the same size.
\color{black}
The linear model and the neural network show systematically similar performance, with an average mean squared error $\text{MSE}\approx 0.01$ and coefficient of determination $\text{R}^{2}\approx 0.99$. The random forest has a slightly lower performance, with average $\text{MSE}=0.12$ and $\text{R}^{2}=0.85$.

We calculate the new feature importance measures discussed in this work: $\widehat{\nu}_j$ (Eq.~\ref{eq:nuhat1}), $\widehat{\nu}_j^{\text{GCMR}}$ (Eq.~\ref{eq:nuhat1} with Alogrithm~\ref{algoGauss}), $\widehat\tau^{\prime}_j$ (Eq.~\ref{eq:nuhat1} with Alogrithm~\ref{algoGauss}),  $\widehat\tau^{\prime, \text{GCMR}}_j$ (Eq.~\ref{eq:tauiprimeLoco} with Alogrithm~\ref{algoGauss}), $\widehat\nu _{j}^\text{GKnock}$ (Eq.~\ref{eq:nuhat1} with the design in Section~\ref{sec:GKnock}), $\widehat\tau^{\text{ALE}}_j$ (Eq.~\ref{eq:tauALEcond}) and $\widehat\kappa^{\text{ALE}}_j$ (Eq.~\ref{eq:kappa}). The local effects needed to compute $\widehat\tau^{\text{ALE}}_j$ and $\widehat\kappa^{\text{ALE}}_j$ are extracted from a suitable \textsc{Matlab} implementation of the corresponding ALE plots R-package. We perform $50$ replicates of the experiments and report the average values.  

\begin{figure}[t]
	\centering
	\begin{subfigure}{0.49\textwidth}   
		\centering 
\includegraphics[width=\textwidth]{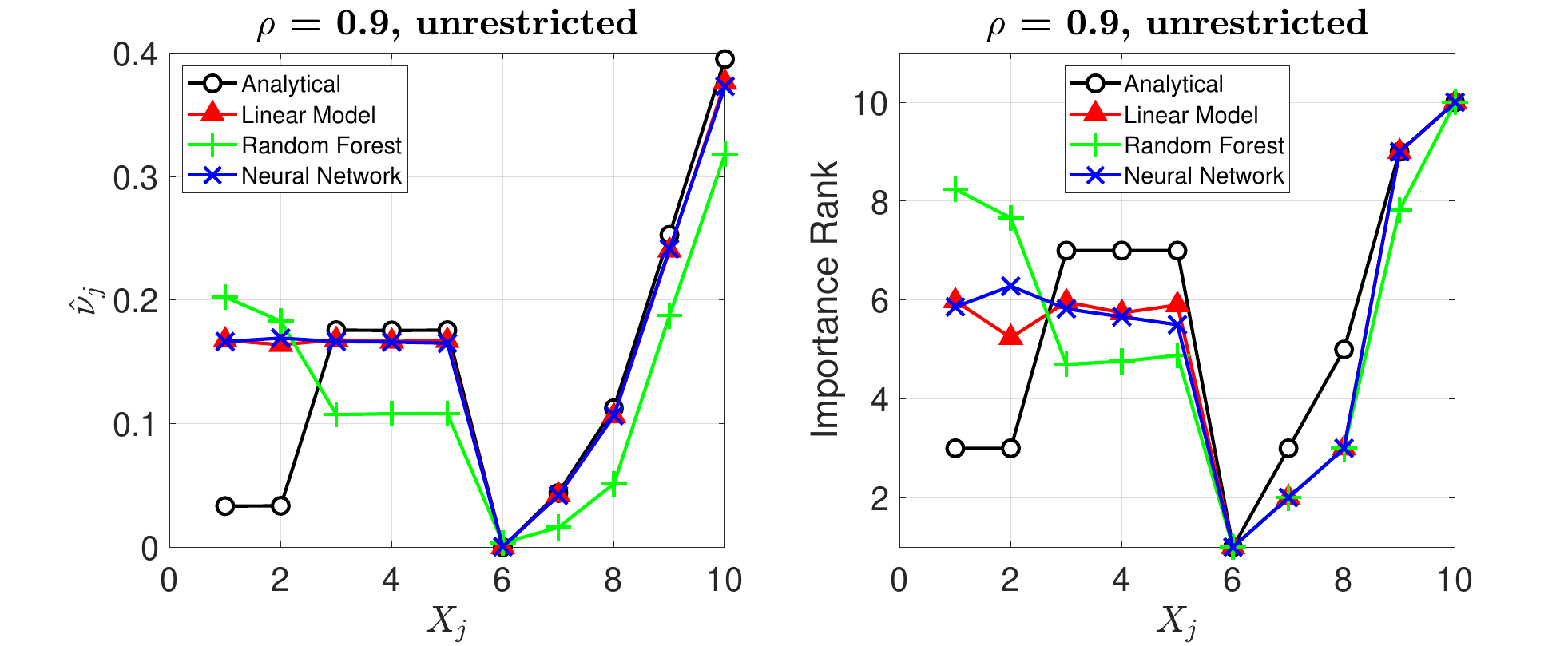}
		\subcaption{$\widehat{\nu}_{j}$ ranking without restrictions.} 
	\end{subfigure}\\
	\begin{subfigure}{0.49\textwidth}   
		\centering 
		\includegraphics[width=0.99\textwidth]{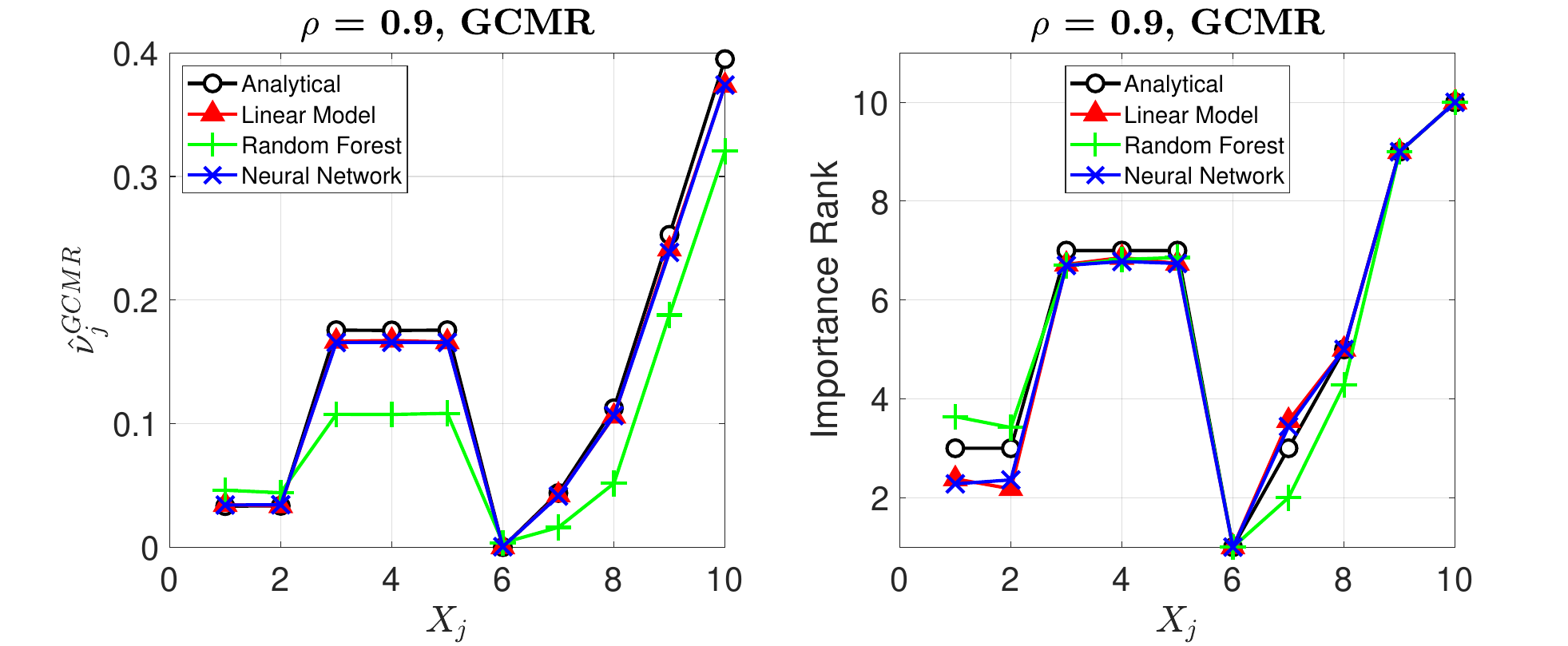}
		\subcaption{$\widehat{\nu}^\text{GCMR}_{j}$ ranking.}
	\end{subfigure}
	\begin{subfigure}{0.49\textwidth}
		\centering
		\includegraphics[width=\textwidth]{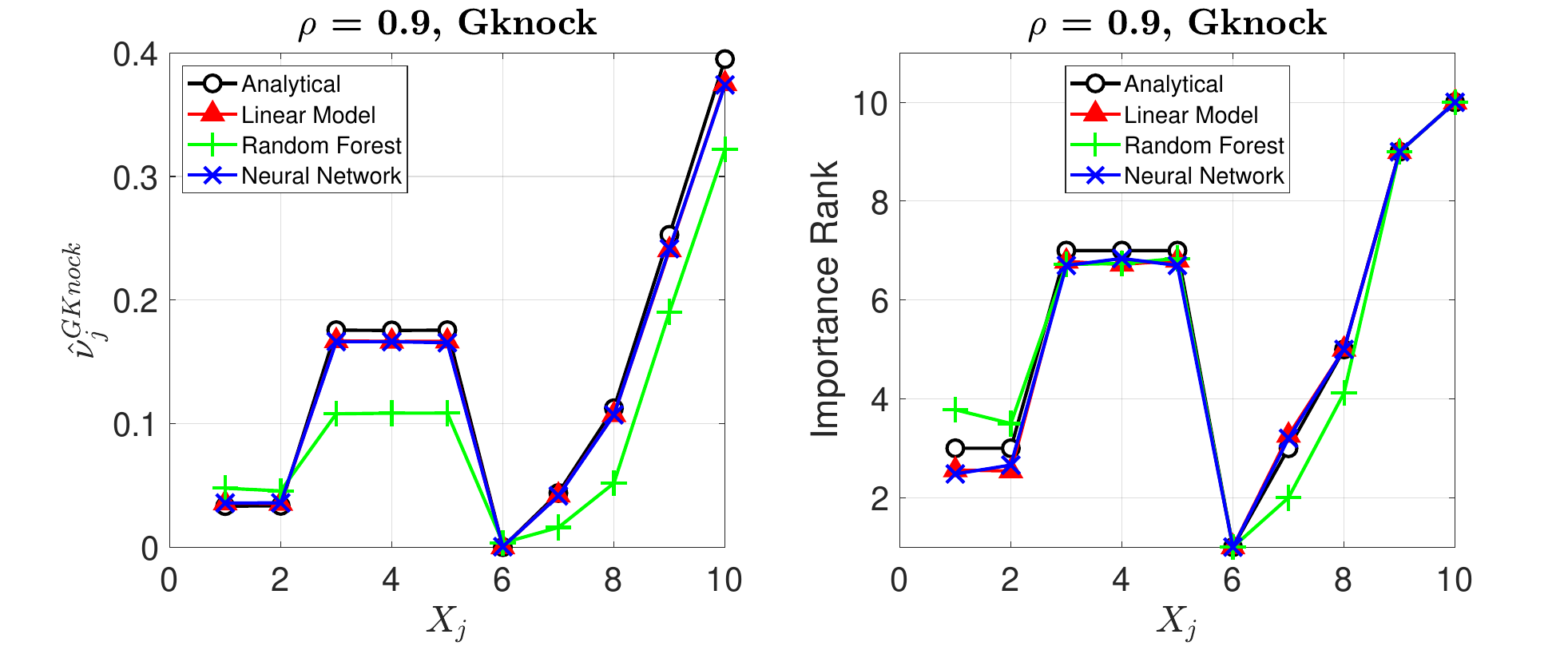}
		\subcaption{$\widehat{\protect\nu}_{j}^{\text{GKnock}}$ ranking.}    
	\end{subfigure}
	\caption{Comparison between estimated feature importance and rankings using $\widehat\nu _{j}$ (the baseline approach), $\widehat\nu_{j}^{\text{GCMR}}$ (ours) and $\widehat\nu_{j}^{\text{GKnock}}$ (also ours) for the Hooker test case. The symbol $\circ$ indicates the theoretical rank of the features. Black is ground truth. Each of these lines is the average ranking over 50 replications.}
	\label{fig:hooketTC2}
\end{figure}

We start with permutation-based importance measures. With $\rho_{X_1,X_2}=0$, we record a correspondence between the ground truth and the estimated values of the variable importance measures for all features and all models. (We omit further discussion for the sake of space.)
However, when $\rho_{X_1,X_2}=0.9$, $\widehat\nu _{j}$ produces different results than $\widehat\nu _{j}^\text{GCMR}$ and $\widehat\nu _{j}^\text{GKnock}$ (Figure \ref{fig:hooketTC2}). Based on the theory, in fact, we expect $\widehat\nu _{j}^\text{GCMR}$ and $\widehat\nu _{j}^\text{GKnock}$ to produce estimates that approximate the ground truth value of the total indices (or their double), as they imply new points generated respecting the involved distributions.  

The left panels of Figure \ref{fig:hooketTC2} report, respectively, the estimates of $\widehat{\nu}_j $, $\widehat{\nu}_j^{\text{GCMR}}$ and $\widehat\tau^{\prime, \text{GCMR}}_j$ on the vertical axis; the right panels corresponds to the importance-ranking (as the information displayed in \cite{Hooker2021}).
\footnote{The Importance Rank equals d-Rank+1. In our case, it is $d=10$ and therefore the most important feature (whose rank is 1) has an Importance Rank of 10.} 
The symbols$\circ $, $\triangle $,  $+$ and $\times $ denote the feature rankings for the original model (theoretical ranking), LM, RF, and NN, respectively. With unrestricted permutations, $\widehat\nu_1$ and $\widehat\nu_2$  differ from the ground truth. They remain close to the independent case for LM and NN, while their importance increases for RF. This is also evident in the ranking shown in the right panel of Figure \ref{fig:hooketTC2}(a).
\color{black}
Now, let's discuss the application of permutation restrictions. Figure \ref{fig:hooketTC2} (lower left) shows that using Gaussian conditional model reliance in Algorithm \ref{algoGauss}, the values of  $\widehat\nu_j^{\text{GCMR}}$ approach the ground truth. For LM and NN models, the estimates nearly match the ground truth, while RF models show greater variability. Figure \ref{fig:hooketTC2} (lower right) indicates that the Gaussian-Knockoff strategy also produces results close to the ground truth for NN and LM, with RF exhibiting more variability.
\color{black}
Note that calculating variable importance measures with permutation restrictions yields lower values than without restrictions. This results from Proposition \ref{prop:nutau} and Corollary \ref{cor:estimates}. Permutation restrictions yield new points $\widehat{g}\left( x_{n,j}^{\prime },\mathbf{x}_{-j}^{n}\right)$ closer to the original points $\widehat{g}\left( 
\mathbf{x}^{n}\right)$, which, for a continuous model, implies smaller squared differences in Equation \eqref{prop:nutau}, and therefore, lower estimates. 
\begin{figure}[t]
\centering
\begin{subfigure}[b]{0.49\textwidth}
            \centering            \includegraphics[width=\textwidth]{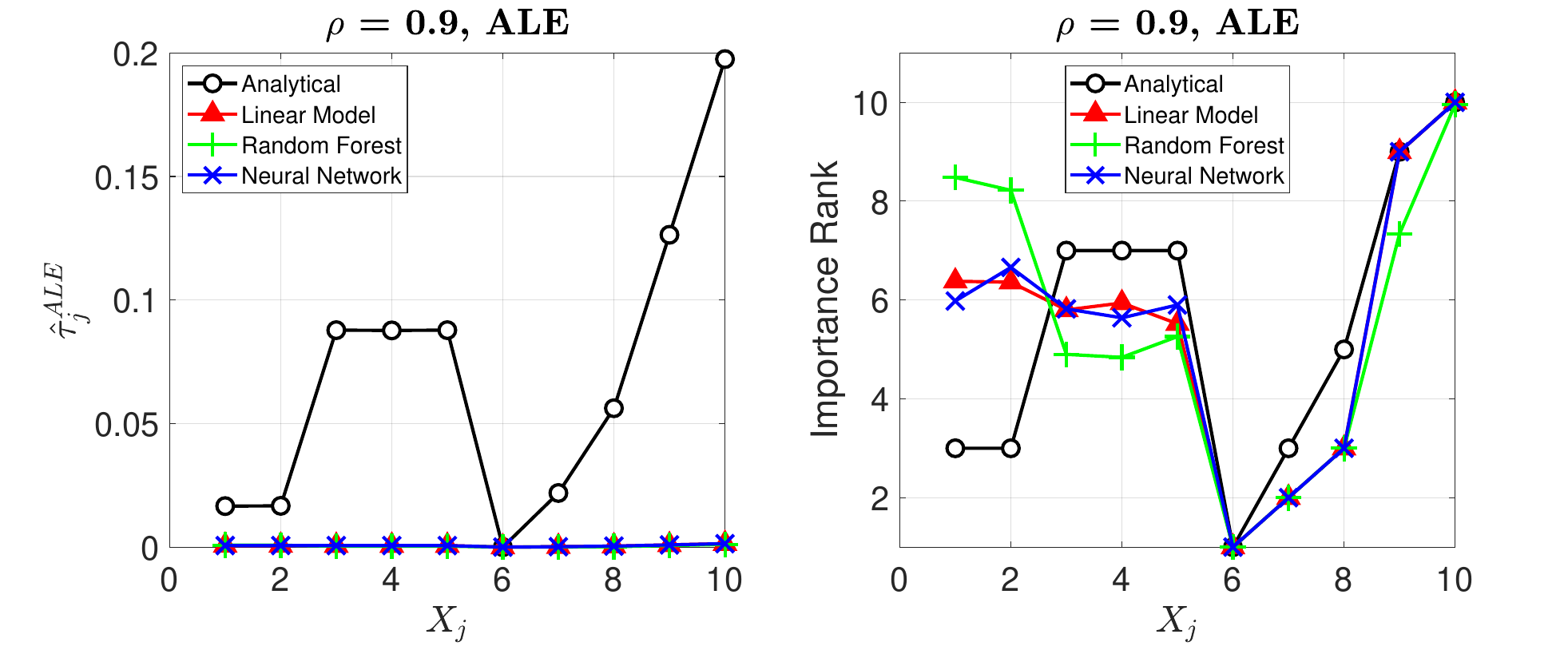}
            \subcaption{Ranking based on $\widehat{\protect\tau}_{j}^\text{ALE}$.}    
        \end{subfigure}
\begin{subfigure}[b]{0.49\textwidth}  
            \centering             \includegraphics[width=\textwidth]{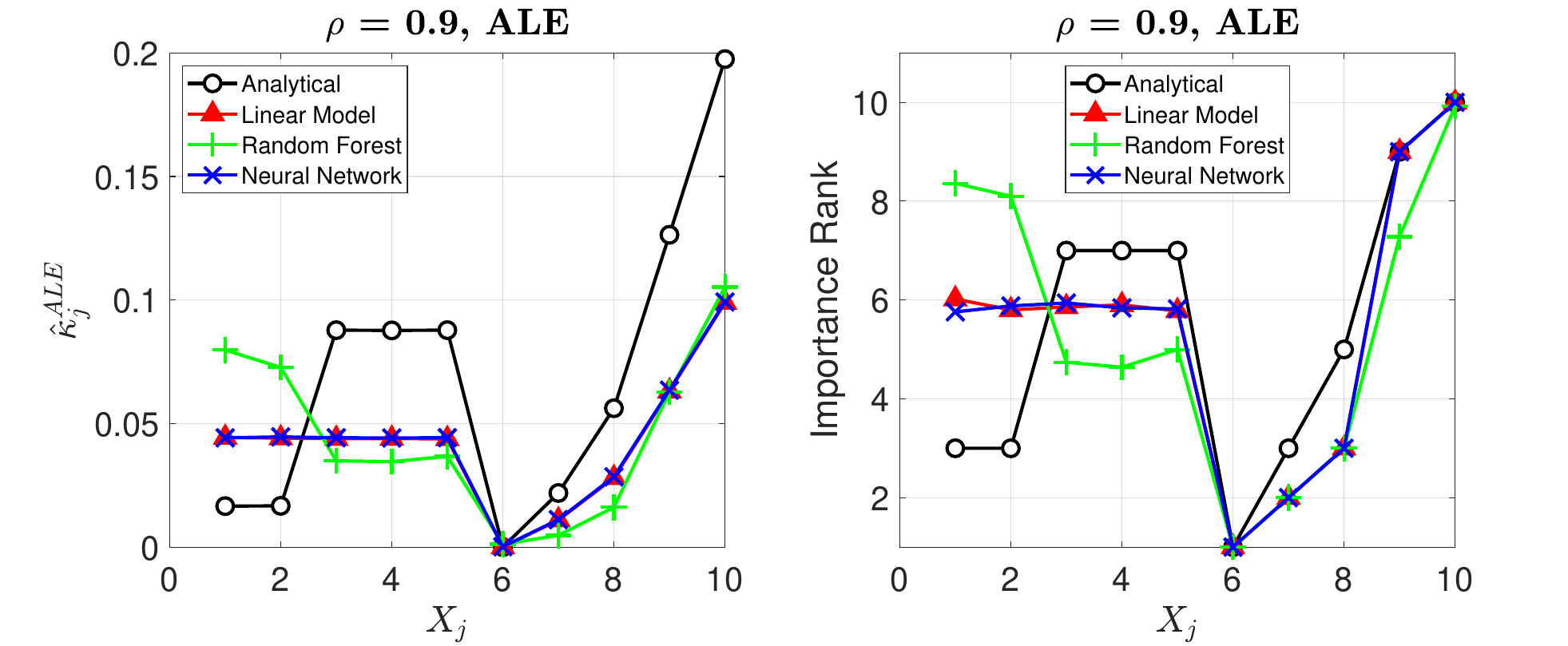}
            \subcaption{Ranking based on $\widehat{\kappa}_{j}^\text{ALE}$.}   
\end{subfigure}
\caption{Estimates (left subpanels) and ranking (right subpanels) of $\widehat{\protect\tau}_{j}^\text{ALE}$ (left) and $\widehat\kappa _{j}^\text{ALE}$ (right) for the Hooker test case. }
\label{fig:hooketTC1}
\end{figure}
Figure \ref{fig:hooketTC1} reports results for the ALE-plot-based indices. The numerical values of $\widehat{\protect\tau}_{j}^\text{ALE}$ (left plot) are far from the values of the ground-truth, and so is the ranking they induce. Besides, $\widehat\kappa _{j}^\text{ALE} $ estimates attribute to $X_1$ and $X_2$ the same importance as $X_3$, $X_4$, and $X_5$ as if they were uncorrelated from the remaining features. The ranking is then consistent with the case in which features are independent or correlated. 
\paragraph{4.2: Boston Housing.}\label{sec:Boston}

The \textit{Boston Housing} dataset \citep{Harrison1978}
is widely used as a reference for
machine learning studies. It has been recorded in 1978 by the U.S Census Service, with 13 features and 506 entries per feature. The target ($Y$) is the value of owner-occupied houses. The target mean $\mathbb{E}[Y]$ and variance $\mathbb{V}[Y]$ are approximately equal to $22.53$ and $84.30$, respectively.

We trained several ML models on this dataset, with the data split into 80\% training and 20\% testing and selected the three best-performing models for our experiments: a linear regression with pairwise interaction terms, an artificial neural network with two seven-neuron layers, and a random forest. The models exhibit the following testing performance: model coefficients of determination ($R^2$) of about $0.92$, $0.85$ and $0.80$, respectively, root mean squared errors of $2.71$, $12$, and $4.5$, respectively, and mean absolute deviations of about $1.82$, $2.70$ and $2.4$, respectively. 

\begin{figure}[htbp]
    \centering
    \includegraphics[width=0.5\textwidth]{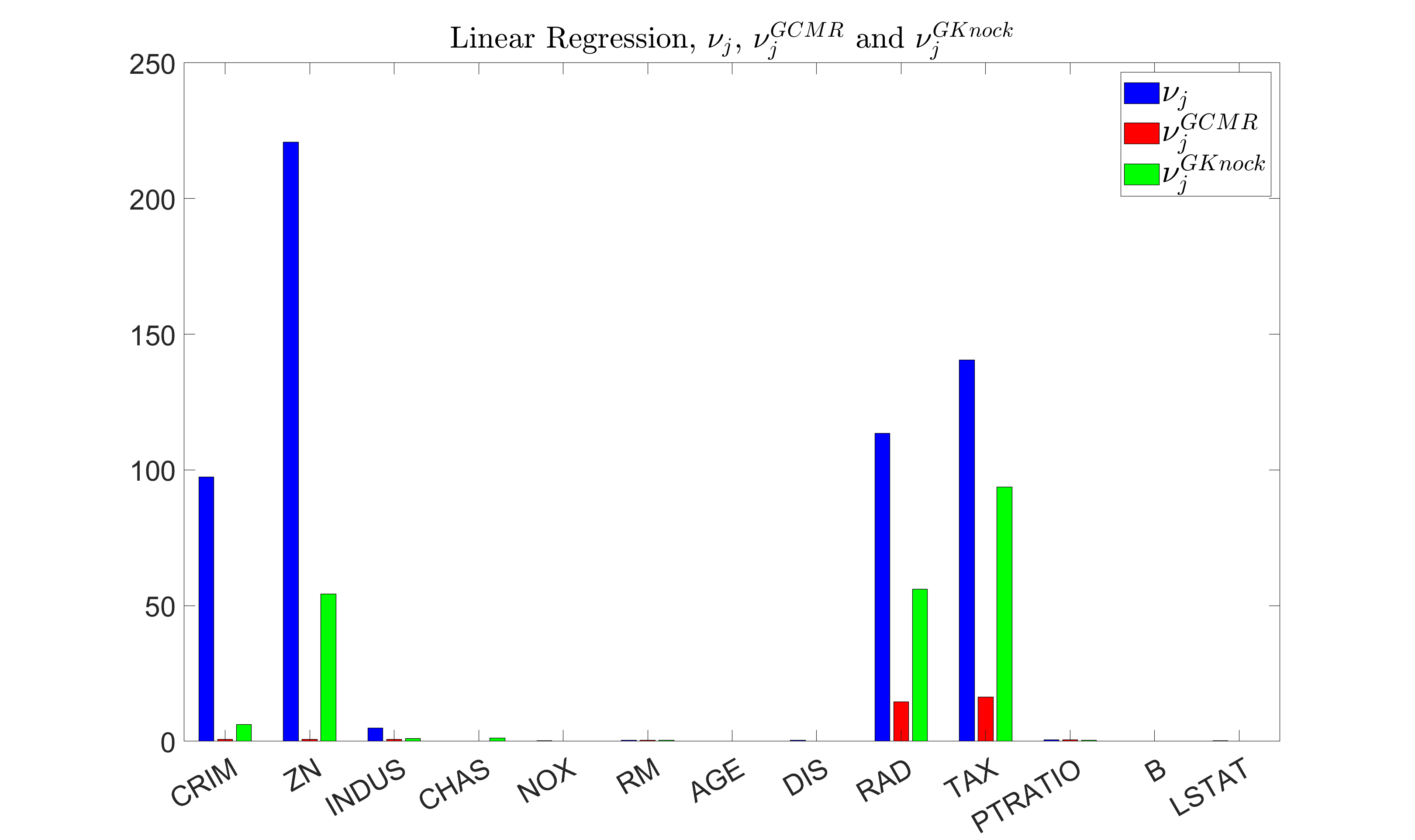}
    \caption{Feature importance measures for the Linear Regression, $\nu_j$, $\nu^\text{GCMR}$ and $\nu^\text{GKnock}$.}
    \label{fig:nutauprimeLinReg}
\end{figure}

\begin{figure}[htbp]
    \centering
    \includegraphics[width=.45\textwidth]{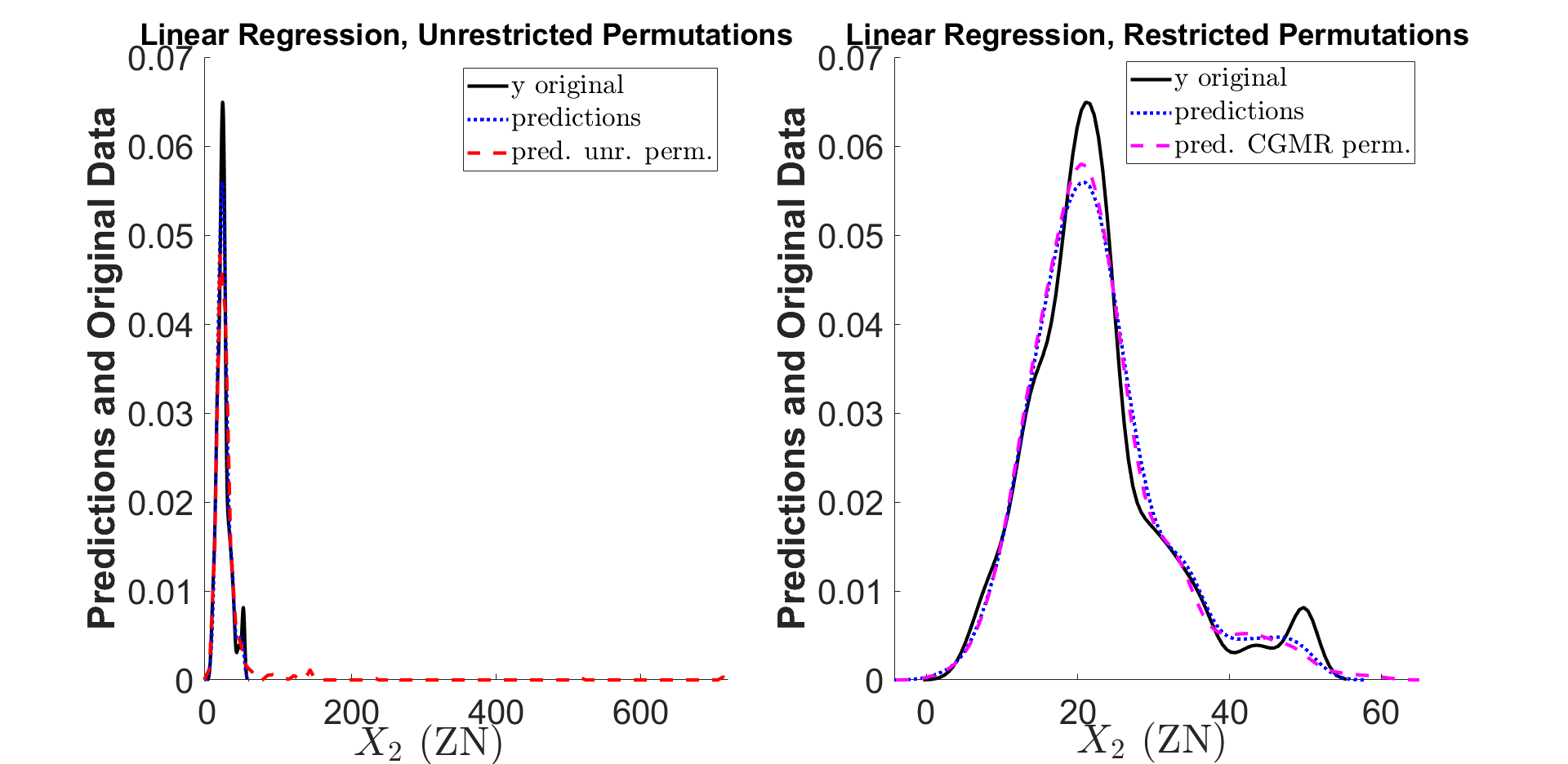}
    \caption{Linear Regression: prediction densities with free permutations of $X_{2}$ (left) and with permutations restricted by GCMR (right, our approach).}
    \label{fig:LinRegX2hist}
\end{figure}

We start with results regarding Breiman's feature importance measures with unrestricted permutations, $\nu_j$, and with restricted permutations $\nu_j^{\text{GCMR}}$, $\tau_j^{\text{GKnock}}$, when the model is the linear regression (Figure \ref{fig:nutauprimeLinReg}).
 Without permutation restriction, for the linear regression, the most important feature would be ZN, followed by TAX, RAD and CRIM, with the remaining features playing a very minor role. 
However, this importance ranking is affected by extrapolation errors, as it is revealed by restricting permutations. Consider features ZN and CRIM. Without restrictions, we record $\nu_{ZN}\approx  220$ and $\nu_{ZN}\approx 95$ (first two blue bars in Figure \ref{fig:nutauprimeLinReg}). After restrictions with GCMR, these values plummet to $\nu^{\text{GCMR}}_{ZN}\approx 0.72$ and $\nu^{\text{GCMR}}_{CRIM}\approx 0.65$. After permutation restrictions, RAD and TAX become the most important variables, regaining a similar position as for the neural network. The green bars in Figure~\ref{fig:nutauprimeLinReg} reveal that the GKnock strategy deflates the feature importance measures, but to a lesser extent than GCMR. (For instance, $\nu_{ZN}$ remains at a high  $\nu^{\text{GKnock}}_{ZN}=54$ even after permutation restriction with GKnock).
We investigate this aspect further considering the model predictions with and without permutations of ZN (Figure~ \ref{fig:LinRegX2hist}).
Comparing the density of the predictions with and without permutation restrictions (dashed red line and dotted blue line, respectively, in Figure \ref{fig:LinRegX2hist}) shows that the heavy and fat tail associated with unrestricted permutations is curtailed when permutations are restricted by GCMR.

\begin{figure}[t]
\centering
\begin{subfigure}[b]{0.49\textwidth}
            \centering  \includegraphics[width=\textwidth]{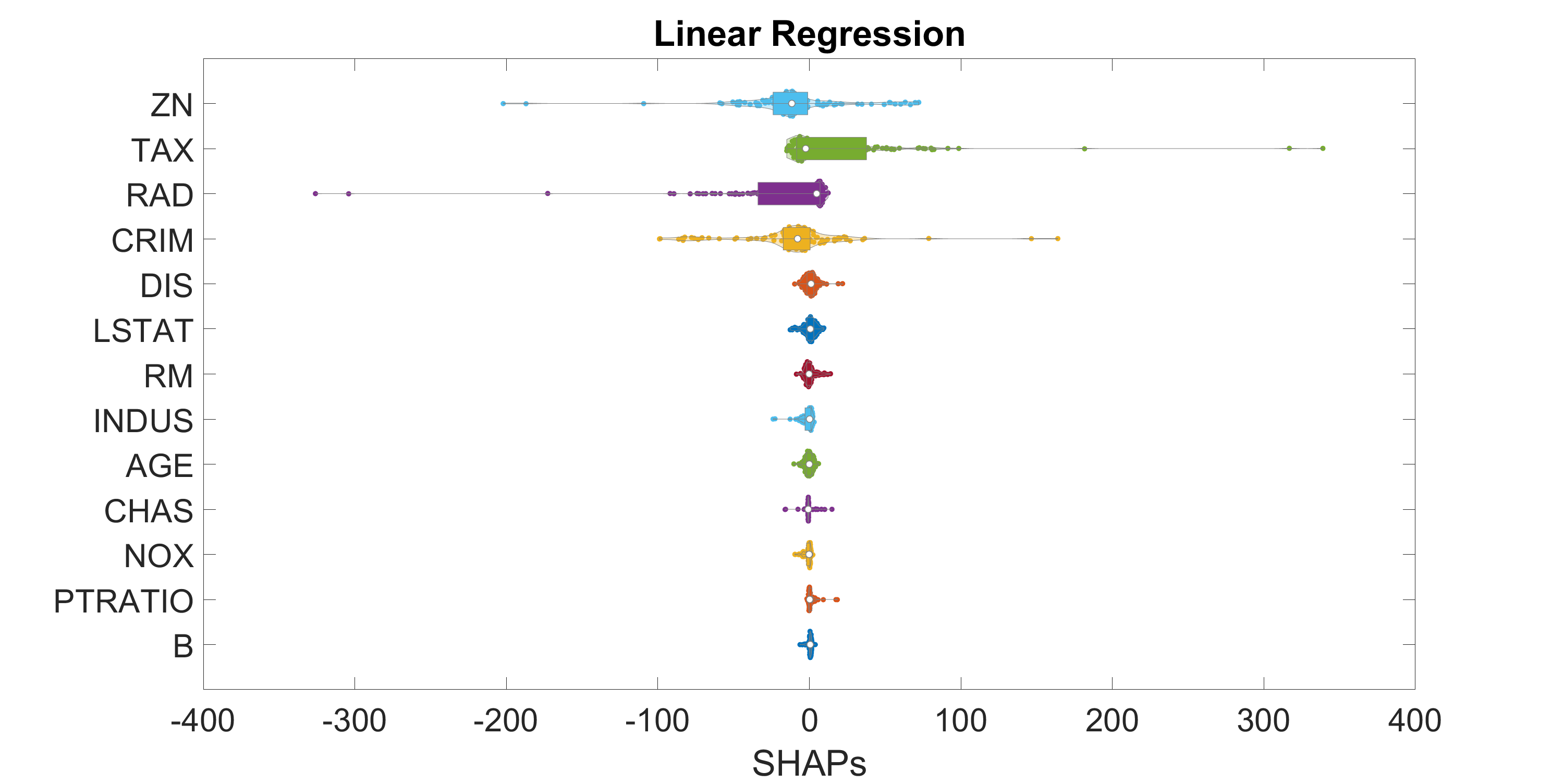}
            \subcaption{\footnotesize SHAP estimates.}    
            \label{fig:MainBostHousShnapsLinRegTornado}
        \end{subfigure}
\begin{subfigure}[b]{0.49\textwidth}  
            \centering 
            \includegraphics[width=\textwidth]{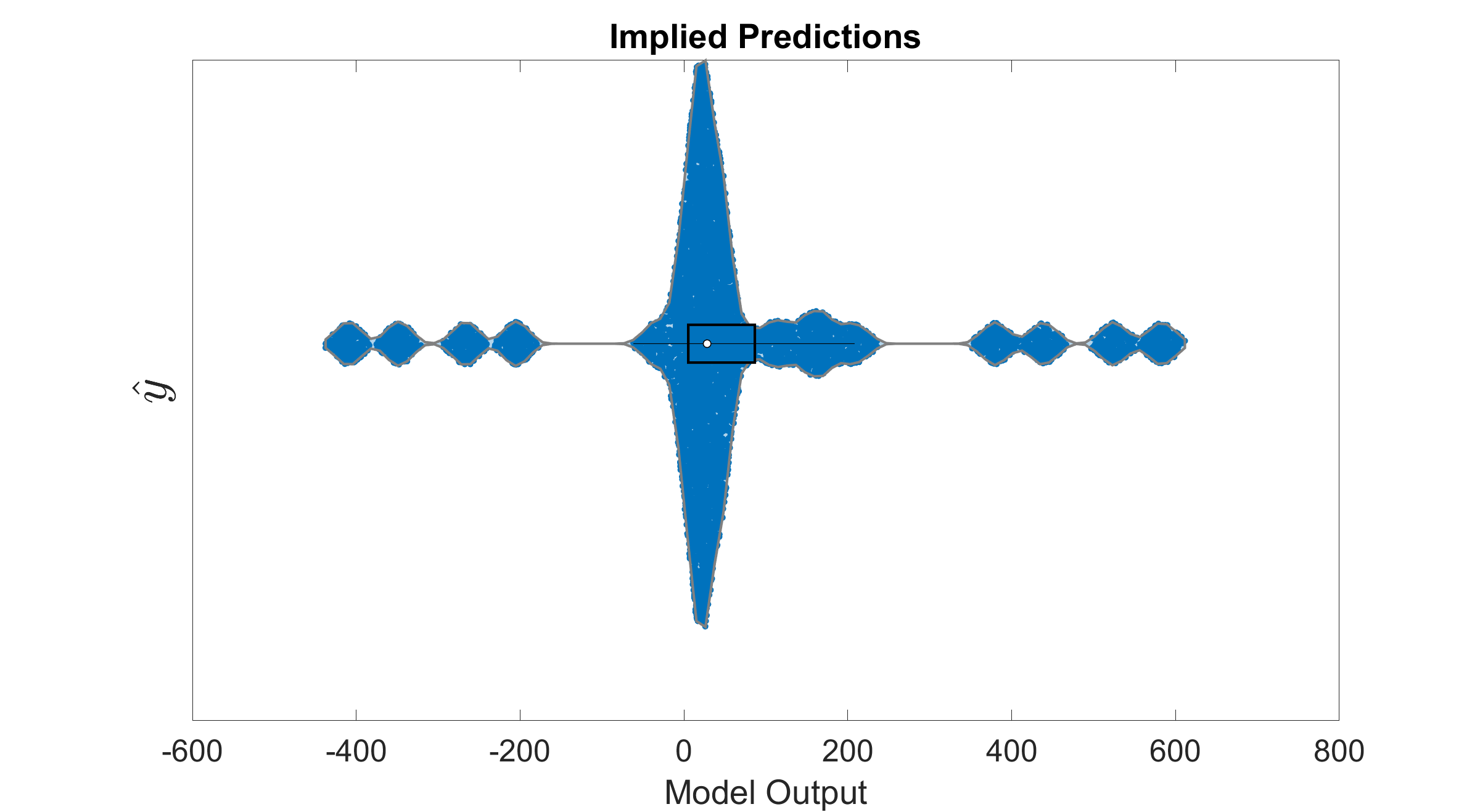}
            \subcaption{\footnotesize Predictions used for calculating SHAPs of TAX.}   
\label{fig:SchnapsExtrapol_hountedhouse}
\end{subfigure}
\caption{SHAP results for the Boston Housing case based on the trained linear regression. Panel (a) shows the SHAP values based on the linear regression model. Panel (b) shows the   Predictions associated with calculating the SHAPs of TAX.}
\label{fig:SHAPLM}
\end{figure}
We then compare these results with an analysis conducted with the SHAPs \citep{Lundberg2020}.
Panel (a) of Figure~\ref{fig:SHAPLM} shows TAX and RAD as the most important features, followed by CRIM, ZN, and INDUS, with NOX, B, and CHAS playing a minor role. 
Notably, TAX values range from slightly below zero to 448, and RAD values from about -400 to just below 10. These values are much larger than the prediction being explained. For example, a SHAP value of 448 for TAX is recorded for house 63, which has a market value of 8.8 and a predicted value of 6.0. 
Panel (b) of Figure~\ref{fig:SHAPLM} shows all predictions used to calculate the SHAP for house 63 for a better understanding. The graph shows several predictions (horizontal axis) with values far outside the original scale, including an implausible prediction of -1000 for House 63. Such predictions would imply receiving a very high sum of money to reside in House 63, a somewhat suspicious insight. Similar issues are seen in RAD (we omit the results for space sake). These results indicate that the SHAP design forces extrapolation, making the SHAP values unreliable.

We analyzed in a similar way the results for the predictions of the neural network and the random forest to study the effects of extrapolations. Overall, the analysis showed a low impact of extrapolations for the neural network and the random forest. The detailed analysis is reported in the full paper.
We also used the methodology to analyze the application of machine learning tools for the analysis of the 
\color{black}
\begin{figure}[htbp]
    \centering
    \begin{subfigure}[b]{0.49\textwidth}
        \includegraphics[width=.9\textwidth]{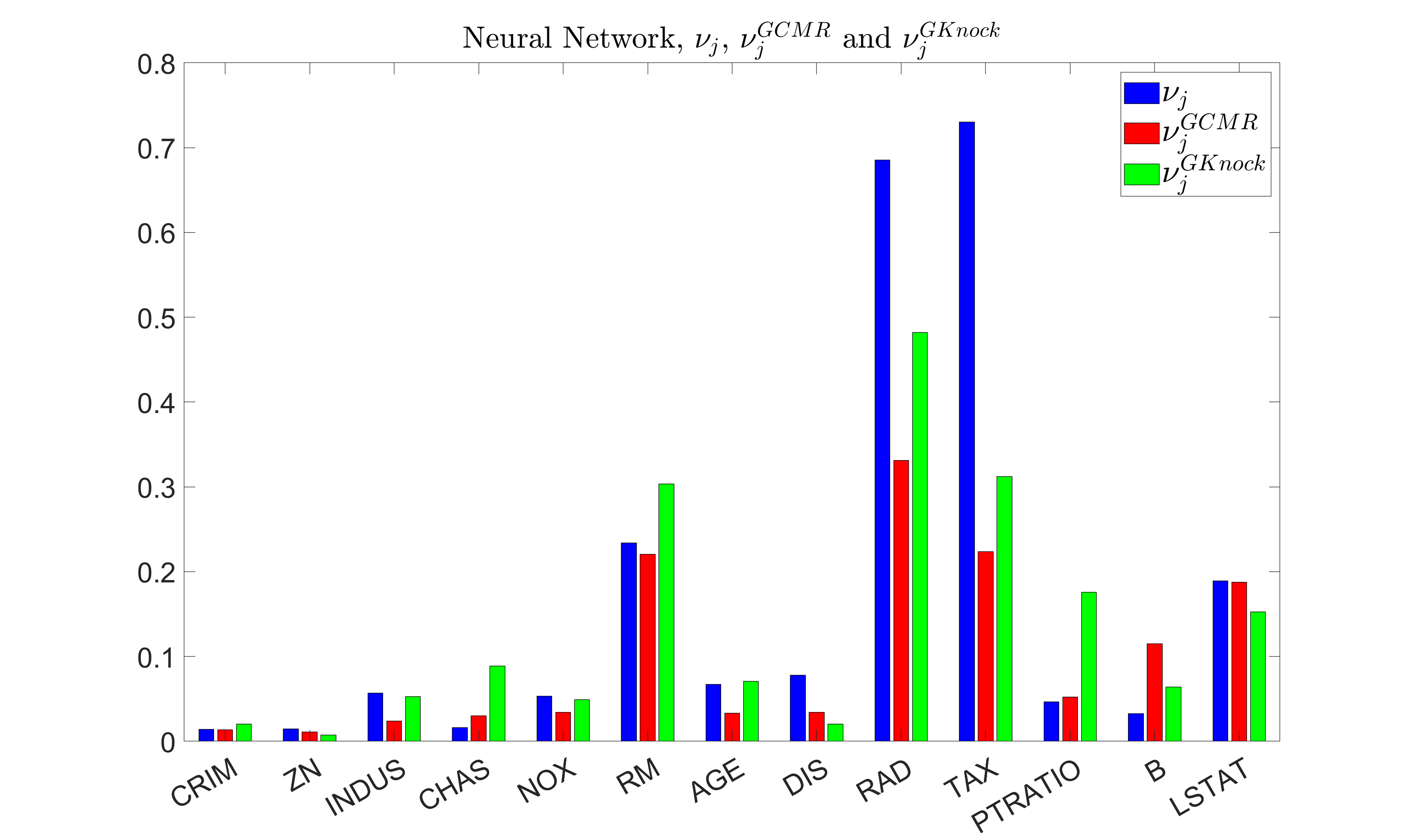}
        \subcaption{Neural Network.}
    \end{subfigure}
    \begin{subfigure}[b]{0.49\textwidth}
        \includegraphics[width=.9\textwidth]{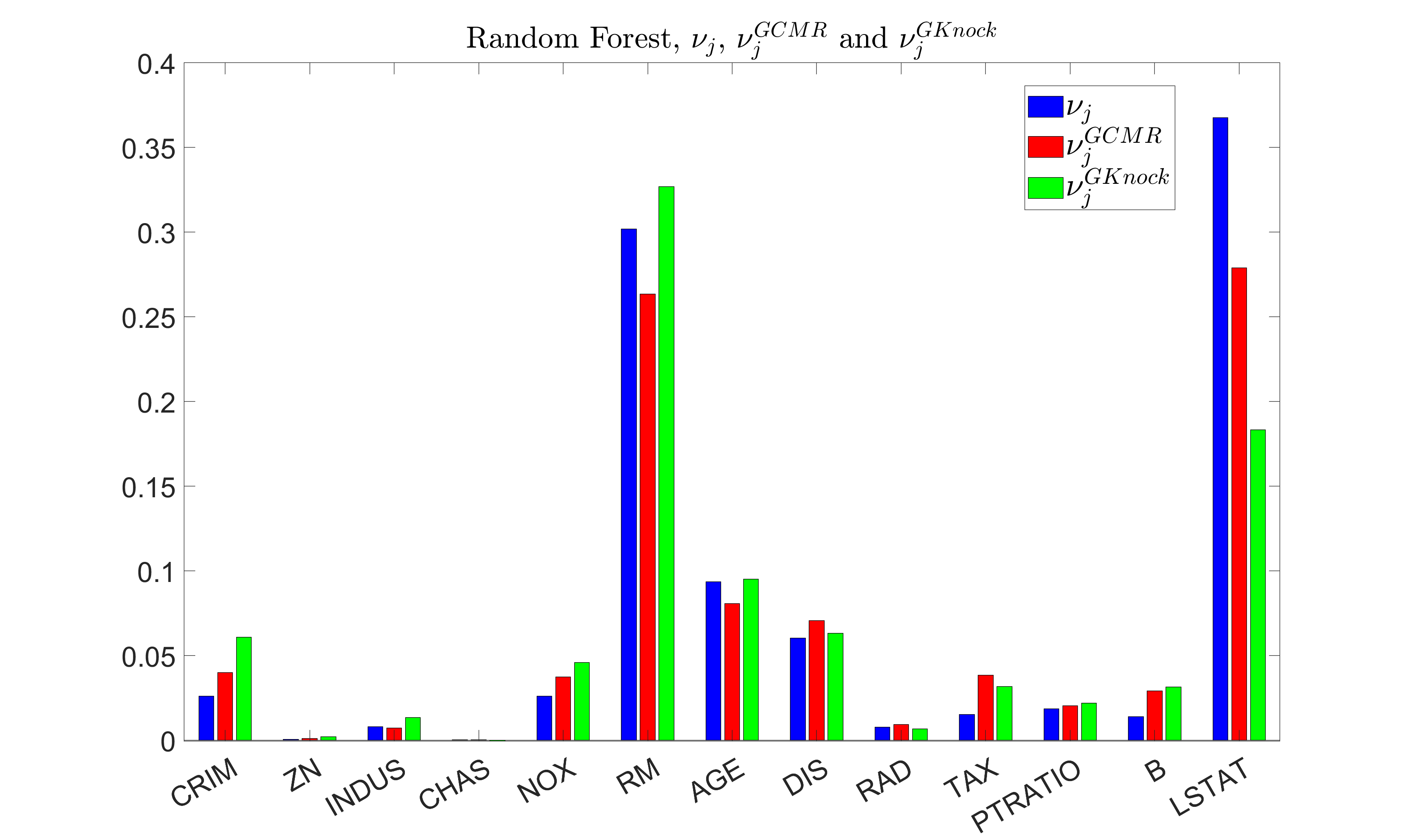}
    sub\caption{Random Forest.}
    \end{subfigure}
    \caption{Permutation based importance measures $\nu_j$, and $\nu_j^{\text{GCMR}}$ (ours), $\nu_j^{\text{GKnock}}$ (also ours) for the neural network.}
\label{fig:MainBostHousGCMRKNockoffsBreiman_NN}
\end{figure}
For the neural network there is an overall agreement among the values of $\nu_j$ (right, blue bars), $\nu_j^{\text{GCMR}}$ (center, red bars) and $\nu_j^{\text{GCMR}}$ that the most important features are RAD ($X_{10}$) and TAX ($X_{11}$), followed by RM ($X_{6}$) and LSTAT ($X_{13}$) with the remaining features having low relevance. The bars associated with TAX and RAD show that their importance is inflated if we do not restrict permutations. To illustrate, $\nu_{10}$ is about 300\% higher than $\nu_{10}^{\text{GCMR}}$ and about 200\% higher than $\nu_{10}^{\text{GKnock}}$. 

For the random forest, the bar charts show that the values of the feature importance measures are much less affected by extrapolation errors than the ones of the other models. This conclusion is also obtained by inspecting the resulting prediction plots. For this model, the most important variables are LSTAT and RM, followed by AGE, CRIM, DIS and TAX. The remaining features play a minor role. 

\color{black}
\paragraph{4.3: The name-ethnicity dataset.}
We used the approach to determine the feature importance measures for two machine learning models fitted to the \textit{name-ethinicity} dataset introduced in \cite{JainEnRu2022}, restricting attention to importance measures that embed permutation restrictions in their definition. The dataset at our disposal contains $13,043,270$ names of voters in selected US States. It is a primary dataset for name-based ethnicity classification studies, which in turn are key tools in determining algorithmic fairness. Results show that the new feature importance measures with permutation restrictions provide reliable information on the feature relevance. The approach also allows us to transparently address the presence of extrapolation via the proposed density comparison plots. Full details on the analysis can be found in the complete paper.
\section{Final Remarks}\label{conclusion}
This work contributes to the ongoing research debate on extrapolation errors in calculating feature importance measures. We conducted an in-depth investigation comparing existing strategies and developing new ones based on alternative designs. For Breiman's variable importance measures, we have proposed a permutation strategy based on conditional model reliance after a Gaussian transformation of the data. This algorithm generates new points that lie close to the original ones. We have provided a theoretical guarantee assuming the statistical dependence among features can be modeled via a Gaussian copula. This strategy yields indices that measure the unique contribution of a feature, excluding the contributions of all other features. Additionally, we have explored generating new points via Knockoffs after a Gaussian transformation, with a theoretical guarantee for this procedure as well.

Within the ALE plot design, we proposed two feature importance measures directly associated with ALE effects. These new indices impose no additional cost for analysts using ALE plots and help avoid false negatives, although their values are sensitive to the choice of the ALE plot grid.

We also studied the relationship of these indices with total effects. We found that under a quadratic loss function, permutation-based importance measures without restrictions are equal to twice  $\Psi_{\text{DLOCO}}$, while with restrictions, they become twice the classical total indices.

Numerical experiments demonstrate that the new permutation strategies (GCMR and GKnock) effectively reduce the inflated importance indices caused by extrapolation errors. However, the $\tau_j^\text{ALE}$ indices remain affected by extrapolation issues.

\bibliographystyle{informs2014}
\bibliography{mybib}

\end{document}